\documentclass[%
 reprint,
 amsmath,amssymb,
 aps,
]{revtex4-1}

\usepackage{graphicx}
\usepackage{dcolumn}
\usepackage{bm}

\usepackage{algorithm}
\usepackage{algorithmic}

\begin{document}

\preprint{APS/123-QED}

\title{Comment on ``Machine learning conservation laws from differential equations"}

\author{Michael F. Zimmer}
 \email{zim@neomath.com}

\date{\today}

\begin{abstract}
The paper \citep{ZLiu-2022} by Liu, Madhavan, and Tegmark sought to use machine learning methods to elicit known conservation laws
for several systems.
However, in their example of a damped 1D harmonic oscillator they made seven serious errors, causing both their method and result to be incorrect.
In this Comment, those errors are reviewed.
\end{abstract}

\maketitle




\subsection*{Review of {\em Liu, et al.}}

In Sec.~III.C of their paper \citep{ZLiu-2022}, they analyzed a damped 1D harmonic oscillator.
With the natural frequency ($\omega_0$) equal to $1$, 
their equation of motion for the position $x$ and momentum $p$ was (cf. Eq.~9 in \citep{ZLiu-2022})
\begin{equation}
\frac{d}{dt}
\begin{pmatrix}
x \\
p 
\end{pmatrix}
 = 
\begin{pmatrix}
p \\
 -x - \gamma p 
\end{pmatrix} \, ,
\label{eqn:EoM}
\end{equation}
and their solution was (cf. Eq.~10 in \citep{ZLiu-2022})
\begin{equation}
\begin{pmatrix}
x (t) \\
p (t)
\end{pmatrix}
 = 
\begin{pmatrix}
e^{-\gamma t} \cos (t + \varphi) \\
e^{-\gamma t} \sin (t + \varphi) 
\end{pmatrix} \, .
\label{eqn:app_critique_soln}
\end{equation}
They then defined $z = e^{ (-\gamma + i)t + i\varphi }$, and used it to define a constant $H' = \varphi$, which they referred to as $H_1$ in Table 1.
Adding intermediate steps, they computed it as
\begin{subequations}
\begin{align}
H_1 
& = \frac{i}{2\gamma} \log \left[ e^{ -2i\gamma\varphi } \right] \\
& = \frac{i}{2\gamma} \left[ -\gamma \log \frac{z}{\bar{z}} - i\log (z\bar{z}) \right] \\
& = -i \log ( e^{i\theta} ) + \log \sqrt{ x^2 + p^2 } / \gamma \\
& = \arctan \left( p/x \right) + \log \sqrt{ x^2 + p^2 } / \gamma \label{eqn:H1-last} \, ,
\end{align}
\label{eqn:H1}
\end{subequations}
where ``$\log$'' refers to the natural logarithm, $\tan \theta = p/x$, 
and $\arctan$ is understood to be evaluated with atan2 \citep{wiki_atan2} with a result on $(-\pi, \pi]$.
(Note they again used the variable name $H_1$ in the text of Sec.~III.C.1, this time to refer to the baseline value $\sqrt{x^2 + p^2}$.)

The errors in these equations will first be summarized, and then will be compared to the equations obtained earlier by Zimmer \citep{MFZ-FJet-arxiv-v2}.

\subsection*{Summary of Errors}

{\bf Errors \#1,2}:
Their solutions $x(t), p(t)$ in Eq.~\ref{eqn:app_critique_soln} do not satisfy the equation of motion (Eq.~\ref{eqn:EoM}).  In particular, they should note
\begin{align*}
\frac{d}{dt} \left[ e^{-\gamma t} \cos (t + \varphi) \right] \neq e^{-\gamma t} \sin (t + \varphi) \, .
\end{align*}
Also, when their $p(t)$ is substituted into Eq.~\ref{eqn:EoM}, it would require ``$x = -x$".
Their errors can be understood by noting
\begin{align*}
& \frac{d}{dt} \cos(t) \neq \sin(t) \\
& \frac{d}{dt} \left[ f(t) g(t) \right] \neq f \frac{dg}{dt} \, ,
\end{align*}
where $f,g$ are two arbitrary functions (cf. Leibniz product rule).

{\bf Error \#3}:
The third error is that $\gamma$ is off by a factor of $2$: either $2\gamma$ should appear in Eq.~\ref{eqn:EoM}, 
or all subsequent instances of $\gamma$ should be replaced by $\gamma/2$. The former change will be assumed.

{\bf Error \#4}:
The fourth error is related to the absence of the pseudo-frequency ($\omega$) due to damping; this is different from the natural frequency ($\omega_0$).
Their $x$ should instead be written as
\begin{subequations}
\begin{align}
& x(t) = A e^{-\gamma t} \cos (\omega t + \varphi) \label{subeqn:corrected_x} \\
& \omega = \sqrt{ \omega_0^2 - \gamma^2} \nonumber \, ,
\end{align}
\label{subeqn:corrected_all}
\end{subequations}
where $A$ is a constant \cite{Rainville-1983,MITOCW_DEs}. 
Note that in Eq.~\ref{eqn:app_critique_soln} they wrote $e^{-\gamma t} \cos ( t + \varphi)$ for $x(t)$, which lacks an $\omega$.
Since they already implicitly set $\omega_0=1$ in Eq.~\ref{eqn:EoM}, if they also set $\omega=1$, then it must be that $\gamma=0$ (i.e., the undamped case).
Also, $\omega$ carries a $\gamma$-dependence, so its absence would have an effect on the comparative plots made at different values of $\gamma$ in their Fig.~4.

Using the corrected version (Eq.~\ref{subeqn:corrected_x}) for $x$ (along with $p=dx/dt$), their variable $z = x + ip$ {\em should} appear as
\begin{align*}
z 
 = & Ae^{-\gamma t} \cos ( \omega t + \varphi) \\
& + i \left[ -\gamma Ae^{-\gamma t} \cos ( \omega t + \varphi) - \omega A e^{-\gamma t} \sin ( \omega t + \varphi) \right] \\
 = & Ae^{-\gamma t} \left[ \cos ( \omega t + \varphi) - i \sin ( \omega t + \varphi + \beta) \right]  \, ,
\end{align*}
with $\sin \beta = \gamma$ and $\cos \beta = \omega$.
Because of the phase angle $\beta$, it's now the case that $z \neq Ae^{ -\gamma t - i(\omega t + \varphi) }$.
Also, since their method was built on using a simple exponential form for $z$, 
it now appears they have no path forward for finding a new expression for $H_1$.

{\bf Error \#5}:
The reader should notice that their derivation was based on cosines and sines, and thus was apparently meant for the underdamped case (they never specified which).
First, the reader should recall (see p.621 of \cite{OlverShakiban-2006} or \citep{MITOCW_DEs,Rainville-1983}) that
the three classes of solutions for this differential equation are: 
(1) underdamped ($\gamma < \omega_0$) with solution set $\{ e^{-\gamma t \pm i\omega t} \}$;
(2) overdamped ($\gamma > \omega_0$) with solution set $\{ e^{-\gamma t \pm \zeta t} \}$, where $\zeta = \sqrt{\gamma^2 - \omega_0^2}$;
(3) critically damped ($\gamma = \omega_0$) with solution set $\{ e^{-\gamma t}, te^{-\gamma t} \}$.
Thus, since they set $\omega_0=1$ and since they're apparently examining the underdamped case, they should limit their numerical tests to where $\gamma < 1$.
However, as their fifth error, they also used it for $\gamma=1,10,100$, i.e. critically damped and overdamped cases.
They also evaluated it in the limit $\gamma \rightarrow \infty$.
(All three cases are analyzed by Zimmer in \citep{MFZ-FJetLie-2024}.)

{\bf Error \#6}: 
Their sixth error was incorrectly evaluating $-i \log ( e^{i\theta} )$ as $\arctan \left( p/x \right)$.
It is well known \citep{wiki_logz} that $\log ( e^{i\theta} )$ is a multivalued function, and is normally understood
as existing on countably many Riemann sheets.  Thus, one should write
$-i \log ( e^{i\theta} ) = [ \theta ]  + 2 \pi n$,
where $[ \theta ]$ takes values in $(-\pi,\pi]$, and the integer $n$ is the sheet number, which accounts for values of $\theta$ outside of $(-\pi,\pi]$.
With this change, Eq.~\ref{eqn:H1-last} should instead appear as
\begin{align*}
H_1 
& = [ \theta ] + 2\pi n + \log \sqrt{ x^2 + p^2 } / \gamma \, .
\end{align*}
Using the $x(t),p(t)$ from Eq.~\ref{eqn:app_critique_soln}, they should have computed $[ \theta] = \arctan(p/x) = [t+\varphi]$ and $\log \sqrt{ x^2 + p^2 } / \gamma = -t$, which leads to
\begin{align*}
H_1 
& = ( [ t+ \varphi ] + 2\pi n) - t \\
& = ( t+ \varphi ) - t \\
& = \varphi  \, ,
\end{align*}
as it was constructed to be.
Also, note that if $2\pi n$ had {\em not} been included $H_1$ could not be a constant as $t$ increases, since $\arctan(p/x)$ is bounded and $-t$ continually decreases.  The most notable consequence of this error is that they omitted the branch cut in their Fig.~4, which was needed to indicate the limits of $\theta$
and the presence of the discontinuity along the negative $x$-axis (both were discussed earlier by Zimmer, cf.~ Fig.~1 of \citep{MFZ-FJetLie-2024}).
In their case, they should have identified the discontinuity as
\begin{align*}
H_1 (x,p; \theta=\pi^-) - H_1 (x,p; \theta= (-\pi)^+) = 2\pi \, .
\end{align*}
This will be made visible in Fig.~1 of this Comment.

{\bf Error \#7}: This error concerns the top row of Fig.~4,
which we are told is comprised of plots of $\cos H_1$ for values $\gamma=0, 0.01, 0.1, 1, 10, 100$.
(Note they used $H_1$ as defined in Eq.\ref{eqn:H1-last}.)

{\underline {\em Issue 1}}: The first issue concerns the plot of $\cos H_1$ for  $\gamma=0$.
By inspection of Eq.~\ref{eqn:H1-last}, $H_1$ is singular at $\gamma=0$.
Thus, as $\gamma \rightarrow 0$, it follows $| H_1 | \rightarrow \infty$, 
and $\cos H_1$ is ill-defined.
Hence, $\cos H_1$ cannot be plotted, and so the plot they show for this case must correspond to some other function.

If they had instead defined $\gamma H_1$ as their constant, this issue would not have appeared (cf.~Eq.~\ref{eqn:r_MFZ}).
Also, doing so would have allowed an easy comparison to the energy of the undamped harmonic oscillator as $\gamma \rightarrow 0$.

\begin{figure}
\includegraphics[scale=0.27]{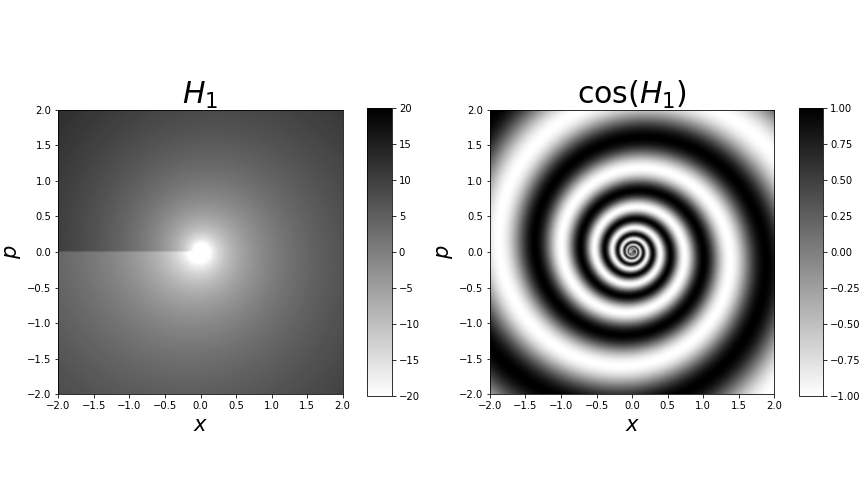}
\caption{ 
Plots of $H_1$ and $\cos H_1$ are displayed together to illustrate how the discontinuity in $H_1$ becomes perfectly obscured when $\cos H_1$ is instead plotted.  In both cases, $H_1$ was computed from Eq.~\ref{eqn:H1-last} with $\gamma=0.1$ and with arctan computed using atan2.
}
\label{fig:H1_cosH1}
\end{figure}

{\underline {\em Issue 2}}: The second issue concerns their choice of plotting $\cos H_1$ rather than $H_1$.  
It's not clear to the author what purpose is served by this non-standard transformation.
It makes it more difficult to compare to the limiting case of $\gamma=0$,
and it also obscures the most interesting feature of $H_1$: its discontinuity (of $2\pi$) along the negative $x$-axis.
Note that because $\cos (H_1 + 2\pi) = \cos(H_1)$, taking the cosine has the side-effect of perfectly hiding this discontinuity.  In addition, $\cos H_1$ has the feature of increasingly tight spirals (as $x,p \rightarrow 0$, or $\gamma \rightarrow 0$), 
which are essentially impossible to model using a standard neural network.
These two features are illustrated in Fig.~\ref{fig:H1_cosH1}, 
where $H_1$ and $\cos H_1$ are plotted side by side.  
Finally, note that if the logical choice of $\gamma H_1$ had instead been used for the constant, then the discontinuity of $2\pi$ would be replaced by $2\pi\gamma$.  Since in general $\cos(H_1 + 2\pi\gamma) \neq \cos(H_1)$, the discontinuity would again be visible.

{\underline {\em Issue 3}}: The third issue is the manifestation of the problem discussed in Error \#5, which was that they took their result derived for $\gamma < 1$ and applied it to plots for $\gamma=1,10,100$.  This means that the plots in the fourth, fifth and sixth columns of the top row of Fig.~4 are invalid.

Finally, given that the first row of Fig.~4 contains problematic or invalid plots, there are obvious questions about the validity of the comparison plots in the third row.

\subsection*{Afterword}

\noindent {\em \underline{Comparison:}}
The reader should note that some of the work summarized in the {\em Review} portion has a counterpart in the earlier work of Zimmer \citep{MFZ-FJet-arxiv-v2}.
For example, Eqs.~3b-3d in this Comment can be matched to similar equations in App.~C of \citep{MFZ-FJet-arxiv-v2}.
(The author has since streamlined his approach in his latest preprint \citep{MFZ-FJetLie-2024}, and now uses different intermediate steps.)

Specifically, the equation of motion used by Zimmer \citep{MFZ-FJetLie-2024} appears as (using $x,p$ in place of $u,v$)
\begin{equation*}
\frac{d}{dt}
\begin{pmatrix}
x \\
p 
\end{pmatrix}
 = 
\begin{pmatrix}
p \\
 - \omega_0^2 x - 2 \gamma p 
\end{pmatrix} \, ,
\end{equation*}
which he used to derive the following constant of motion
\begin{align}
r
& = \log [ \omega^2 x^2 + ( \gamma x + p )^2 ] -  2 \frac{\gamma}{\omega} ( \phi - 2\pi n) \, ,
\label{eqn:r_MFZ} 
\end{align}
where
\begin{align*}
\omega & = \sqrt{ \omega_0^2 - \gamma^2 } \\
\tan \phi & = (\gamma x + p) / (\omega x) \, ,
\end{align*}
where $\phi$ is confined to a $2\pi$-interval, and $n \in \mathbb{Z}$ is the Riemann sheet number.
Note that as $\gamma \rightarrow 0$, the constant $r$ smoothly approaches $\log (2E)$, where $E$ is the energy of the undamped harmonic oscillator.
Finally, all three damping cases for the harmonic oscillator were treated by Zimmer \citep{MFZ-FJetLie-2024}.

\noindent {\em \underline{Their analytical approach:}}
In their treatment of the 1D damped oscillator, they began from exact solutions for $x,p$ and then formed combinations of them to isolate
the parameters of the solution, thereby determining a constant of motion.
Such an approach can only be used, as they presented it, in the undamped case.
To make it work in the damped case, they should have made a variable change ($x \rightarrow w = \gamma x + p$) that Zimmer recognized
(see App.~D.2 of \citep{MFZ-FJetLie-2024}).
However, Liu, et al.~gave no indication they were aware of such a transformation.


\noindent {\em \underline{2D oscillator:}}
Liu, et al.~also analyzed the 2D oscillator (undamped), and made similar omissions regarding Riemann sheets,
as they did in the 1D case.
However, most notable is that they did not express their constant as a function of $x,p$ (i.e., in terms of ``$\tan^{-1}$"); they instead kept it as a function of the solution parameters.  They thus missed the chance to see how such a result could be used as a generalization of angular momentum,
as well as other results (see Secs.~VI E-G in \citep{MFZ-FJetLie-2024}).
Also, they would then go on to to write the following pejorative remarks, with which the author disagrees:
(1) in the caption of Fig.~5 on p.~045307-7, ``an everywhere discontinuous function that is completely useless to physicists";
(2) on p.~045307-6, ``ill-behaved, demonstrating fractal behavior".
It is suggested here that this may be a graphical aliasing effect.

\noindent {\em \underline{Generalized $H_1$:}}
Finally, it's noted that the evaluation of their constant as $H_1 = \varphi$ wasn't as general as it could have been
(disregarding for now the errors already discussed).
Specifically, they should have written $x(t) = A \cos (t + \varphi)$ as their general solution (with $A>0$).
With this, their constant $H_1 = \varphi$ would then be replaced by $H_1 = \varphi + (\log A ) / \gamma$.
This would have been useful in computing the value of $H_1$ along an arbitrary trajectory,
which could then be used in a comparison check to level set values of a contour plot of $H_1$ (cf.~analogous checks done in \citep{MFZ-FJetLie-2024}).



\bibliography{BIB_Comment_Tegmark_2}

\end{document}